\documentclass[sigconf]{acmart}
\usepackage{fancyhdr}

\AtBeginDocument{%
  \providecommand\BibTeX{{%
    \normalfont B\kern-0.5em{\scshape i\kern-0.25em b}\kern-0.8em\TeX}}}

\newlength\savewidth

\author{Jintao Sun}
\affiliation{%
 \department{School of Computer Science and Technology}
  \institution{Beijing Institute of Technology}
  \city{Beijing}
  \country{China}
}
\email{3120215524@bit.edu.cn}

\author{Hao Fei}
\affiliation{%
  \department{School of Computing}
  \institution{National University of Singapore}
  \city{Singapore}
  \country{Singapore}
}
\email{haofei37@nus.edu.sg}

\author{Gangyi Ding}
\affiliation{%
\department{School of Computer Science and Technology}
  \institution{Beijing Institute of Technology}
  \city{Beijing}
  \country{China}
}
\email{dgy@bit.edu.cn}

\author{Zhedong Zheng}
\authornote{Corresponding author.}
\affiliation{%
  \department{Faculty of Science and Technology, and Institute of Collaborative Innovation}
  \institution{University of Macau}
  \city{Macau}
  \country{China}
}
\email{zhedongzheng@um.edu.mo}
\copyrightyear{2025}
\acmYear{2025}
\setcopyright{acmlicensed}
\acmConference[WWW '25]{Proceedings of the ACM Web Conference 2025}{April 28-May 2, 2025}{Sydney, NSW, Australia}
\acmBooktitle{Proceedings of the ACM Web Conference 2025 (WWW '25), April 28-May 2, 2025, Sydney, NSW, Australia}
\acmDOI{10.1145/3696410.3714788}
\acmISBN{979-8-4007-1274-6/25/04}

\def\eg{\emph{e.g.}} 
\def\ie{\emph{i.e.}} 
\def\etal{\emph{et~al.}}

\usepackage {bbding}
\begin{document}

\title{From Data Deluge to Data Curation: A Filtering-WoRA Paradigm \\ for Efficient Text-based Person Search}

%\renewcommand{\shortauthors}{Trovato and Tobin, \etal}

\begin{abstract}
  %With the advent of Vision-Language Pretraining (VLP), the performance of numerous vision-language downstream tasks has been significantly enhanced. Many tasks attempt to bolster model learning capabilities by constructing datasets with a broad array of information.
    %In this paper, we examine a specific task, Text-guided Person Search. 
  %Recent literature has proposed methods involving the construction of datasets with extensively generated data for training. Data generation has been widely deployed in the person search tasks, considering the privacy protection and annotation difficulties.It remains a scentific problem that how diversity and fidelity data could facilitate the subsequential model training% thereby accelerating model training. 
  %Consequently, this enables more effective training with fewer data, enhancing the learning capability of the model. By selectively updating a minimalist ensemble of model parameters, %While pretraining with a large volume of generated data can enhance model performance,% to retain data surpassing a defined threshold of relevance. 
 In text-based person search endeavors, data generation has emerged as a prevailing practice, addressing concerns over privacy preservation and the arduous task of manual annotation. Although the number of synthesized data can be infinite in theory, the scientific conundrum persists that how much generated data optimally fuels subsequent model training. We observe that only a subset of the data in these constructed datasets plays a decisive role. Therefore, we introduce a new Filtering-WoRA paradigm, which contains a filtering algorithm to identify this crucial data subset and WoRA (Weighted Low-Rank Adaptation) learning strategy for light fine-tuning. The filtering algorithm is based on the cross-modality relevance to remove the lots of coarse matching synthesis pairs. As the number of data decreases, we do not need to fine-tune the entire model. Therefore, we propose a WoRA learning strategy to efficiently update a minimal portion of model parameters. WoRA streamlines the learning process, enabling heightened efficiency in extracting knowledge from fewer, yet potent, data instances. Extensive experimentation validates the efficacy of pretraining, where our model achieves advanced and efficient retrieval performance on challenging real-world benchmarks. Notably, on the CUHK-PEDES dataset, we have achieved a competitive mAP of 67.02\% while reducing model training time by 19.82\%.
\end{abstract}





\maketitle

\begin{figure}[h]
  \centering
  \vspace{-1.6em}
    \includegraphics[width=0.98\linewidth]{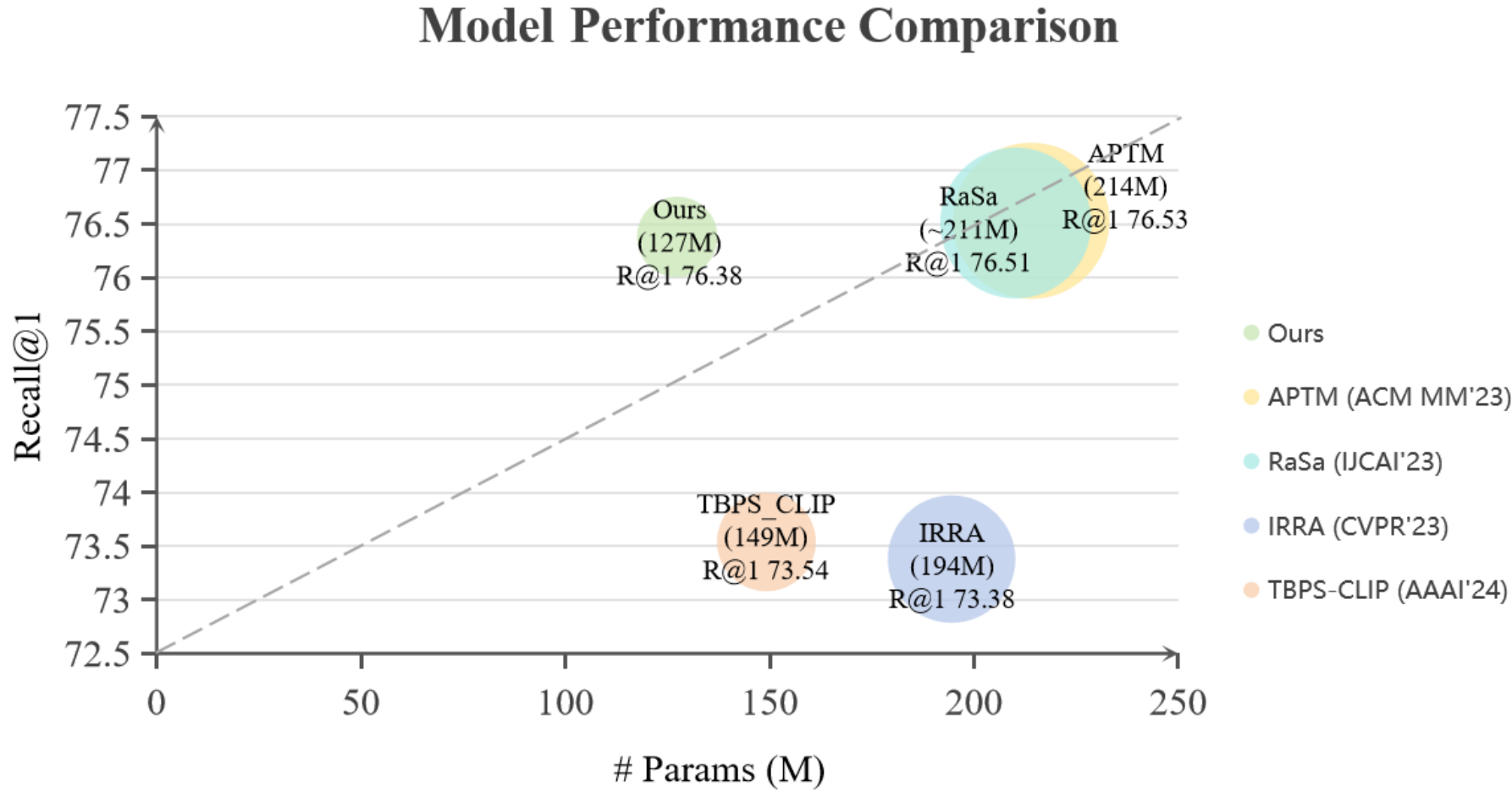}
  % \caption{Comparison between the proposed method and existing approaches in terms of mAP accuracy and the parameter numbers. We observe that our method deploys fewer parameters while achieving a higher mAP than competitive methods, \ie, APTM~\cite{APTM}, IRRA~\cite{jiang2023crossmodal} and TBPS-CLIP~\cite{cao2023empirical}.}
  \vspace{-.15in}
    \caption{Comparison between the proposed method and existing approaches in terms of Recall@1 and the parameter numbers. We observe that our method deploys fewer parameters while achieving a competitive Recall@1, \ie, APTM~\cite{APTM}, RaSa~\cite{Bai_2023}, IRRA~\cite{jiang2023crossmodal} and TBPS-CLIP~\cite{cao2023empirical}.}
  \Description{Comparison between the proposed method.}
  \vspace{-1.6em}
\label{Comparison between}
\end{figure}

\section{Introduction}
%Traditional Image Re-Identification (ReID) involves matching the same entities across different non-overlapping camera views~\cite{Liao_Hu_Zhu_Li_2014,Xiao_Li_Wang_Li_Wang_2016,Zheng_Gong_Xiang_2011}. 
Compared to traditional image-based person search ~\cite{zhu2021dssl,Shao_Zhang_Fang_Lin_Wang_Ding_2022,Wang_Zhu_Xue_Wan_Liu_Wang_Li_2022,CAIBC,book,10091915,9652475}, which seeks to retrieve target individuals from a vast array of footage or images across different locations and times, 
text-based person search locates interested individuals from a pool of candidates based on pedestrian descriptions~\cite{Li_Xiao_Li_Yang_Wang_2017}. Given that pedestrian image queries may not be available, % in some complex or specific scenarios, such as remote roads without cameras or locations where pedestrians are completely obscured,
text-guided person search emerges as an alternative method. 
The key lies in mining the fine-grained information from images and texts, blending the complexity of natural language processing with the subtle nuances of visual recognition, and establishing their correspondence. 
By leveraging the ability to understand complex human descriptions and accurately identify and retrieve images of individuals from a camera system, it can be applied to broad applications in public safety domains such as missing person searches \cite{10531327,10377578}, and rescue operations \cite{9369386}. As the sub-task of vision-language retrieval, text-based person search models usually require extensive data for training, where, however, the number of pedestrian data is limited. 

Most datasets~\cite{Li_Xiao_Li_Zhou_Yue_Wang_2017, zhu2021dssl,Ding_Ding_Shao_Tao_2021,APTM} are constructed from three sources. 
(1) The first source is through sampling from camera footage, accompanied by manual annotations. However, constructing large-scale datasets %suitable for vision-language model training
is often infeasible due to privacy concerns and high costs. 
(2) The second source involves collecting images and short videos from the internet. Despite expanding dataset sizes, the noisy web text and the inconsistent quality of task images are generally sub-optimal for fine-grained vision-language learning. 
(3) Therefore, most researchers~\cite{APTM} resort to leverage the generative models, \eg, GAN~\cite{zheng2017unlabeled} and Diffusion~\cite{9878449,podell2023sdxl}. For instance, APTM~\cite{APTM} has introduced 1.51M image-text pairs generated by Stable Diffusion~\cite{podell2023sdxl}, showcasing the potential of training on a large synthesized dataset.

Despite significant progress in learning from large synthesized datasets, a fundamental challenge persists: \textbf{how can we efficiently extract knowledge when faced with an effectively infinite amount of data, considering the substantial computational costs incurred during training?} 
% Despite notable advances in learning from large synthesized datasets, a core challenge remains: \textbf{how can we efficiently extract knowledge from an effectively infinite data volume, given the high computational costs of training?}
% We observe two points: 
% (1) Prior works~\cite{zheng2017unlabeled,zheng2019joint,APTM} have shown that performance improvements taper off even when copious amounts of additional generated data are supplied. 
% This implies that not all information within massive synthetic datasets is equally valuable; rather, a carefully selected subset, or coreset, may suffice to capture the essence of the training process.
% (2) If we only need to learn coreset data, it is not necessary for models to update the entirety of parameter volume to ensure training precision. 
% This realization opens up the possibility of significantly trimming down both training time and model complexity. In essence, our focus shifts towards exploring strategies that enable heightened accuracy with leaner data and a more compact parameter footprint.
We observe two key points: (1) Previous works~\cite{zheng2017unlabeled,zheng2019joint,APTM} indicate that performance gains diminish even with abundant generated data. This suggests that not all information in massive synthetic datasets is equally useful; instead, a well-chosen subset, or coreset, may be sufficient to capture the essential training information. (2) When learning from a coreset, updating the entire parameter space is unnecessary to achieve training precision. This insight opens the door to significantly reducing both training time and model complexity. Thus, our focus shifts to developing strategies that achieve high accuracy with reduced data and a compact parameter footprint.
%Second, model pre-training and fine-tuning require a substantial number of parameters, thereby continuously increasing the demand for computational power while also significantly impacting training time. 
%To meet practical application needs, models usually need to learn data more quickly and efficiently to fulfill task requirements. 
%Despite end-to-end training with vision-language models and large datasets, 
%Therefore, most researchers resort to pretraining and fine-tuning paradigm. 
%Most state-of-the-art vision-language models contain the same number of parameters during pre-training and fine-tuning as the original models.
%However, for higher accuracy, most of them employ large, deep networks to learn high-dimensional real-value features. Thus, the introduction of pre-training and the learning of more extensive data significantly slows down the model learning speed, especially when dealing with large datasets.
%Additionally, Model performance is compared in Figure ~\ref{Comparison between}. 

%sets three parameters— $\alpha$, $\beta$, and m—to perform parameter learning 
%inspired by the LoRA~\cite{hu2021lora} and DoRA~\cite{liu2024dora} methods,
%, such as BLIP-2~\cite{li2023blip2}, capability to %understand image-text pairs. This involves 

To this end, we propose a new Filtering-WoRA paradigm, which contains a novel two-stage data filtration method aimed at identifying the coreset to enhance model performance and a WoRA (Weighted Low-Rank Adaptation) algorithm to optimize the pre-training and fine-tuning models, enabling training with fewer parameters while maintaining model performance and increasing computational speed. Specifically, our process begins with dataset purification, including the 
synthesized dataset for pre-training and the 
real-world dataset for fine-tuning. To filter out low-quality image-text pairs, \eg, incomplete descriptive details or blurred image details, we leverage the 
off-the-shelf large cross-modality model to 
extract features from both images and texts within the dataset and then calculate the cosine similarity between projected image embeddings and projected text embeddings. 
This process yields a similarity score for each image-text pair, facilitating the selection of high-quality datasets based on our predetermined threshold. 
Subsequently, to reduce model parameters and enhance computational speed, we opt to freeze the weights from pre-training, indirectly training some dense layers in the neural network by optimizing rank decomposition matrices that change during the adaptation process. 
By decomposing pre-training and fine-tuning weights into magnitude and direction, our WoRA method introduces three new dimensions to facilitate the modification of the weight matrix and rank decomposition matrix. 
This approach allows for learning a minimal amount of parameters while simultaneously boosting model performance (see Figure ~\ref{Comparison between}). 
In summary, our contributions are: % as follows:
\begin{itemize}
\vspace{-0.2in}
% \item We introduce a new Filtering-WoRA paradigm for efficient text-based person search, which streamlines learning and improves efficiency through focused data curation and targeted parameter updates. 
% The filtering algorithm targets relevant, high-quality synthesized data by assessing cross-modality relevance, while WoRA (Weighted Low-Rank Adaptation) enables lightweight fine-tuning of a minimal set of model parameters.
\item We propose Filtering-WoRA, a paradigm for efficient text-based person search that streamlines learning through focused data curation and targeted parameter updates. Our filtering algorithm assesses cross-modality relevance to select high-quality synthesized data, while Weighted Low-Rank Adaptation (WoRA) enables lightweight fine-tuning of a minimal set of model parameters.
\item Extensive experiments on three widely-used benchmarks verify that our method could save 19.82\% training time compared with the vanilla baseline, while also achieving competitive 76.37\%, 66.65\%, 67.90\% Recall@1 accuracy on CUHK-PEDES, RSTPReid and ICFG-PEDES, respectively.
\end{itemize}

\vspace{-0.20in}
\section{Related Work}
%item We introduce a two-stage data filtration method to select high-quality image-text pairs for model training.
%\\item We devise a new low-parameter learning strategy, WoRA, which, by designing three parameters, learns only a minimal amount of mo
%del parameters, significantly enhancing model computational speed while ensuring training precision.
% As noted in the introduction, the ``\verb|acmart|'' document class can
% be used to prepare many different kinds of documentation --- a
% dual-anonymous initial submission of a full-length technical paper, a
% two-page SIGGRAPH Emerging Technologies abstract, a ``camera-ready''
% journal article, a SIGCHI Extended Abstract, and more --- all by
% selecting the appropriate {\itshape template style} and {\itshape
%   template parameters}.

% This document will explain the major features of the document
% class. For further information, the {\itshape \LaTeX\ User's Guide} is
% available from
% \url{https://www.acm.org/publications/proceedings-template}.

\noindent\textbf{Vision-Language Pre-training.}
% vision-language pre-training (VLP) aims to learn visual-linguistic alignments from a copious corpus of image-text pairs. The paradigm of "pre-training followed by fine-tuning" stands as a cornerstone in propelling the field of computer vision. This approach involves initially pre-training models on extensive datasets, subsequently fine-tuning them for specific downstream tasks to achieve state-of-the-art performance, such as in visual question answering and image-text retrieval. In this context, the quality of the pre-trained models plays a pivotal role in the complexity of model optimization during the fine-tuning phase and the eventual performance outcomes.
Current VLP research predominantly bifurcates into coarse-grained and fine-grained methodologies. Coarse-grained approaches employ convolutional networks~\cite{Jiang_Misra_Rohrbach_Learned-Miller_Chen_2020,Huang_Zeng_Liu_Fu_Fu_2020,Huang_Zeng_Liu_Fu_Fu_2021,zhang2023structure} or visual Transformers~\cite{Kim_Son_Kim_2021,SwinTransformer,ALBEF,BEITV2,kan2024lfde} to extract and encode holistic image features, thereby constructing vision-language models (VLMs). Techniques such as SOHO~\cite{Huang_Zeng_Liu_Fu_Fu_2021} propose leveraging a Visual Dictionary (VD) for the extraction of comprehensive yet compact image features, facilitating enhanced cross-modal comprehension. ALBEF~\cite{ALBEF} introduces a contrastive loss for aligning image and text representations before their fusion through cross-modal attention, fostering more grounded vision-language representation learning. Additionally, it utilizes momentum distillation to augment learning capabilities from noisy network data. Although these holistic image-focused methods are efficient, their performance is generally outpaced by fine-grained approaches.  Inspired by advancements in the NLP domain, fine-grained methods~\cite{Tan_Bansal_2019,Lu_Batra_Parikh_Lee_2019,Li_Yatskar_Yin_Hsieh_Chang_2019,Li_Duan_Fang_Gong_Jiang_2020,Chen_Li_Yu_ElKholy_Ahmed_Gan_Cheng_Liu_2020,Li_Yin_Li_Zhang_Hu_Zhang_Wang_Hu_Dong_Wei_etal._2020,Gan_Chen_Fu_Zhu_Cheng_Liu_2020} employ pre-trained object detectors~\cite{Ren_He_Girshick_Sun_2017,Anderson_He_Buehler_Teney_Johnson_Gould_Zhang_2018} trained on annotated datasets of common objects, such as COCO~\cite{Lin_Maire_Belongie_Hays_Perona_Ramanan_Dollár_Zitnick_2014} and Visual Genome~\cite{Krishna_Zhu_Groth_Johnson_Hata_Kravitz_Chen_Kalantidis_Li_Shamma_etal._2017}. This enables models to recognize and classify all potential object regions within images, representing them as a collection of object-centered features. For instance, VinVL~\cite{Zhang_Li_Hu_Yang_Zhang_Wang_Choi_Gao_2021} enhances visual representations for V+L tasks and develops an improved object detection model to provide object-centered image representations. However, this object-centered feature representation struggles to capture relationships between multiple objects across different regions, limiting its effectiveness in encoding multi-granularity visual concepts. Another limitation is the inability of the object detector to recognize uncommon objects not present in the training data. Recently, novel approaches have emerged to bridge the learning of object-level and image-level alignments. E2E-VLP~\cite{Xu_Yan_Li_Bi_Huang_Xiao_Huang_2021} employs DETR~\cite{Carion_Massa_Synnaeve_Usunier_Kirillov_Zagoruyko_2020} as the object detection module to enhance detection capabilities. KD-VLP~\cite{Liu_Wu_Tseng_Lal_He_Duan_2022} relies on external object detectors for object knowledge distillation, facilitating cross-modal alignment learning across different semantic layers. OFA~\cite{wang2022ofa} formulates visual-linguistic tasks as a sequence-to-sequence (seq2seq) problem, adhering to instruction-based learning during both pre-training and fine-tuning phases, eliminating the need for extra task-specific layers. Uni-Perceiver~\cite{zhu2021uniperceiver} constructs a unified perception architecture, using a single Transformer and shared parameters for diverse modes and tasks, employing a non-mixed sampling strategy for stable multi-task learning. 
X-VLM~\cite{zeng2022multigrained} and X2-VLM~\cite{zeng2022x} propose an integrated model with a flexible modular architecture to simultaneously learn multi-granularity alignment and localization, achieving the capability to learn infinite visual concepts related to various text descriptions. In this work, we leverage the proficient vision-language pre-trained model to filter noisy data. %enhance performance on the person search task.
% \bfseries dd
% The primary parameter given to the ``\verb|acmart|'' document class is
% the {\itshape template style} which corresponds to the kind of publication
% or SIG publishing the work. This parameter is enclosed in square
% brackets and is a part of the {\verb|documentclass|} command:
% \begin{verbatim}
%   \documentclass[STYLE]{acmart}
% \end{verbatim}

% Journals use one of three template styles. All but three ACM journals
% use the {\verb|acmsmall|} template style:
% \begin{itemize}
% \item {\verb|acmsmall|}: The default journal template style.
% \item {\verb|acmlarge|}: Used by JOCCH and TAP.
% \item {\verb|acmtog|}: Used by TOG.
% \end{itemize}

% The majority of conference proceedings documentation will use the {\verb|acmconf|} template style.
% \begin{itemize}
% \item {\verb|acmconf|}: The default proceedings template style.
% \item{\verb|sigchi|}: Used for SIGCHI conference articles.
% \item{\verb|sigchi-a|}: Used for SIGCHI ``Extended Abstract'' articles.
% \item{\verb|sigplan|}: Used for SIGPLAN conference articles.
% \end{itemize}

% \vspace{1mm}
\noindent\textbf{Text-Image Person Search.}
% \zznote{Please move the reference to the places.}
Based on the challenging task of language-based person search, which is a fine-grained, cross-modal retrieval challenge, a significant number of methodologies have been developed in recent years to tackle this issue. Existing approaches can broadly be classified into two categories: those based on cross-modal attention interaction~\cite{Li_Xiao_Li_Zhou_Yue_Wang_2017,Shao_Zhang_Fang_Lin_Wang_Ding_2022,Shu_Wen_Wu_Chen_Song_Qiao_Ren_Wang_2022,Wang_Zhu_Xue_Wan_Liu_Wang_Li_2022} and those without cross-modal attention interaction~\cite{Chen_Zhang_Lu_Wang_Zheng_2022, Ding_Ding_Shao_Tao_2021,CAIBC,Zheng_Zheng_Garrett_Yang_Xu_Shen_2020,yang2024walkinglargescaleimagetextbenchmark,APTM}. 
% Methods leveraging cross-modal attention interaction facilitate cross-modal correspondences between regions and words or phrases through paired inputs and predict image-text pair matching scores via attention mechanisms. This enriches interaction between modalities, bridging the modality gap at the cost of higher computational complexity. For example, Li \etal~\cite{Li_Xiao_Li_Zhou_Yue_Wang_2017} propose a novel recurrent neural network with gated neural attention to improve cross-modal learning. Shao \etal~\cite{Shao_Zhang_Fang_Lin_Wang_Ding_2022} propose a multimodal shared dictionary (MSD) to reconstruct visual and textual features while using a set of shared and learnable archetypes as queries. In order to improve the performance of person search, different and semantically consistent features are extracted for the two modes in the feature space with uniform granularity. 
Methods leveraging cross-modal attention interactions facilitate correspondences between image regions and textual phrases by pairing inputs and predicting image-text matching scores through attention mechanisms. This enhances interaction between modalities, effectively bridging the modality gap, albeit at the cost of increased computational complexity. For instance, Li \etal~\cite{Li_Xiao_Li_Zhou_Yue_Wang_2017} propose a recurrent neural network with gated neural attention to enhance cross-modal learning. Shao \etal~\cite{Shao_Zhang_Fang_Lin_Wang_Ding_2022} introduce a multimodal shared dictionary (MSD) to reconstruct visual and textual features, employing shared, learnable archetypes as queries. To improve person search performance, features for both modalities are extracted in the feature space with uniform granularity, ensuring semantic consistency.
Conversely, methods without cross-modal attention interaction, through the construction of diverse model structures and objective functions, align representations of the two modalities within a shared feature space. These lightweight models, not reliant on complex cross-modal interactions, are computationally more efficient and have even achieved better results than their attention-based counterparts. Zheng~\etal~\cite{Zheng_Zheng_Garrett_Yang_Xu_Shen_2020} build an end-to-end dual-path convolutional network to learn image and text representations to take full advantage of supervision capabilities. %In this paper, we propose a new dual-path local alignment network architecture for extracting visual and text local representations, and then propose a multi-stage cross-modal matching strategy that eliminates modal gaps from three feature levels: low, local, and global. 
However, all the person search methods mentioned above %learn real-value features 
fine-tune the entire network for high accuracy, which inherently slows down the process. %Therefore, 
In contrast, we propose a method based on cross-modal feature extraction for rapid candidate selection and data ranking scores. This approach aims to mitigate the impact of low-quality text-image pairs while reducing model computational parameters to maintain high efficiency in processing.

\noindent\textbf{Data-Centric Learning.}
With the surge in popularity of large language models, an increasing demand for vast datasets for model training has emerged~\cite{CLIP,zeng2022multigrained,zeng2022x,APTM}. However, open-source datasets constructed for training models in real-world scenarios, such as the MALS dataset~\cite{APTM}, may encounter issues like incorrect text descriptions, poor image or text quality, and insufficient feature matching between image-text pairs, all of which can adversely affect model training performance. As the size of datasets expands, it is observed that their quality does not invariably scale in tandem~\cite{zheng2017unlabeled}. Frequently, a subset of high-caliber data can attain or even exceed the utility of a voluminous but qualitatively inferior dataset. This phenomenon underscores the paramount importance of meticulously curating high-quality datasets, thereby highlighting the necessity for efficiency and precision in dataset construction. For instance, data selection methods~\cite{radenovic2023filtering,wu2024visionlanguage,huang2024data} aim to identify and train only with the most relevant and informative examples, discarding irrelevant or redundant data. This leads to more efficient learning and improved model performance, especially when dealing with large datasets. Active learning seeks to reduce labeling costs by selecting the most informative instances for annotation. Data cleaning and preprocessing techniques aim to remove noise, errors, and inconsistencies from the data, making it more suitable for learning. Coreset selection~\cite{li2022blip,li2023blip2} focuses on identifying a small subset of data points (a core set) sufficient for training a model that performs well across the entire dataset. By selecting a representative subset, coreset selection can reduce the computational costs of training and improve model generalization. Our approach advocates for a coreset method aimed at enhancing model performance, proposing the deploy of the off-the-shelf visual-language models to segment and filter the dataset for text and image pair matching scores, thereby obtaining a coreset dataset for effective training.

% As datasets grow larger, their quality does not necessarily improve proportionally. Often, only a fraction of high-quality data can achieve or even surpass the effectiveness of a more extensive, yet lower-quality dataset, making the efficient construction of high-quality datasets crucial.
\begin{figure*}[ht]
\vspace{-1.3in}
  \centering
  \includegraphics[width=\linewidth]{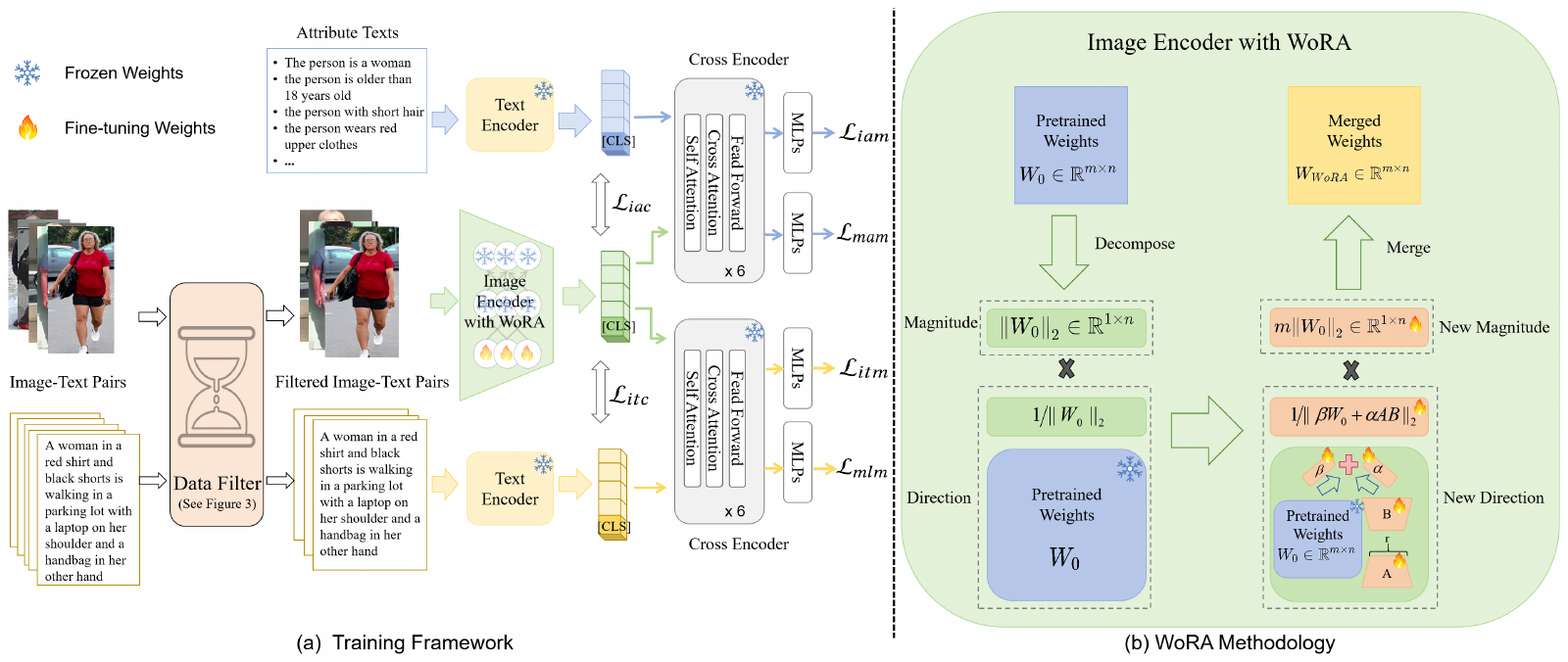}
    \vspace{-1.4in}
  \caption{An overview of our framework. %The left half of the picture 
  (a) shows the flow chart of the entire training pipeline. 
  We first obtain the filtered training image-text pairs. Then we augment the text as attribute texts according to keywords. We extract the corresponding features through image encoder, text encoder and cross encoder. There are six loss objectives of both text-image and attribute-image matching tasks.
  %To the right, We present
  (b) is an in-depth illustration of WoRA methodology, meticulously applied within the context of an image encoder. The model is updated by fine-tuning the decomposition of the pre-trained weights into amplitude and direction components and updating both components using LoRA~\cite{hu2021lora} while adding the $\alpha$ and $\beta$ learnable parameters. Since the image encoder consumes most GPU memory and time. In practice, we mainly apply the WoRA on the image encoder.}  
  \Description{An overview of our Architecture.}
\label{arte}
\vspace{-.1in}
\end{figure*}

\section{Methodology}
%In this section, we first re-introduce the baseline structure, then outline our data filtering strategy, followed by a detailed presentation of our WoRA method for efficiently reducing the model parameters. The brief architecture of our Filtering-WoRA is shown in Figure ~\ref{arte}.

\subsection{Baseline Revisit}
We do not pursue the network contribution in this work, but focus on the training efficiency. Our method can be adopted to most existing works. Without loss of generality, we apply the widely-used baseline, APTM~\cite{APTM}, to simplify the illustration as well as a fair comparison with other methods. In particular, 
the framework comprises three encoders, \ie, image encoder, text encoder, and cross encoder, along with two MLPs-based headers.
The entire training process contains two phases, \ie, pre-training on the synthesized dataset and fine-tuning on the downstream datasets. As shown in Figure~\ref{arte}, the $\left [ \mathrm{CLS}  \right ] $ embedding represents the aggregated feature of the image / text from the image encoder and the text encoder respectively. %For textual encoding, , which first tokenizes the input text into $N+1$ tokens and processes them through the first 6 layers of Bert to produce text embeddings. 
%The $\left [ \mathrm{CLS}  \right ] $ token embodies the entire text representation. 
The cross encoder integrates image and text representations for prediction tasks, leveraging the latter 6 layers of Bert to process the previously obtained text and image embeddings, thereby discerning their semantic relationship.
%In this study, 
We adopt two types of loss functions to bolster alignment constraints, tailored for both image-text and image-attribute associations. The image-text functions encompass Image-Text Contrastive Learning (ITC), Image-Text Matching Learning (ITM), and Masked Language Modeling (MLM), while  Attribute Prompt Learning contains Image-Attribute Contrastive Learning (IAC) loss, Image-Attribute Matching Learning (IAM) Loss, and Masked Attribute Modeling (MAM) Loss.
The overall APL loss is $\mathcal{L} _{APL}=\frac{1}{3}  \left ( \mathcal{L} _{iac}+\mathcal{L} _{iam}+\mathcal{L} _{mam} \right ) $, and
%Similarly, we can get Image-Text Matching Learning (ITM) and Masked Language Modeling (MLM), 
the full pre-training loss is formulated as: $\mathcal{L} _{total}=  \mathcal{L} _{itc}+\mathcal{L} _{itm}+\mathcal{L} _{mlm} +\eta \mathcal{L} _{APL}$, where $\eta$ is empirically set as 0.8 following~\cite{APTM}. %We re-introduce our baseline and loss function because it inspires us in desinging the WoRA method.

\subsection{Data Filtering}
\begin{figure}[t]
  \centering
  \includegraphics[width=\linewidth]{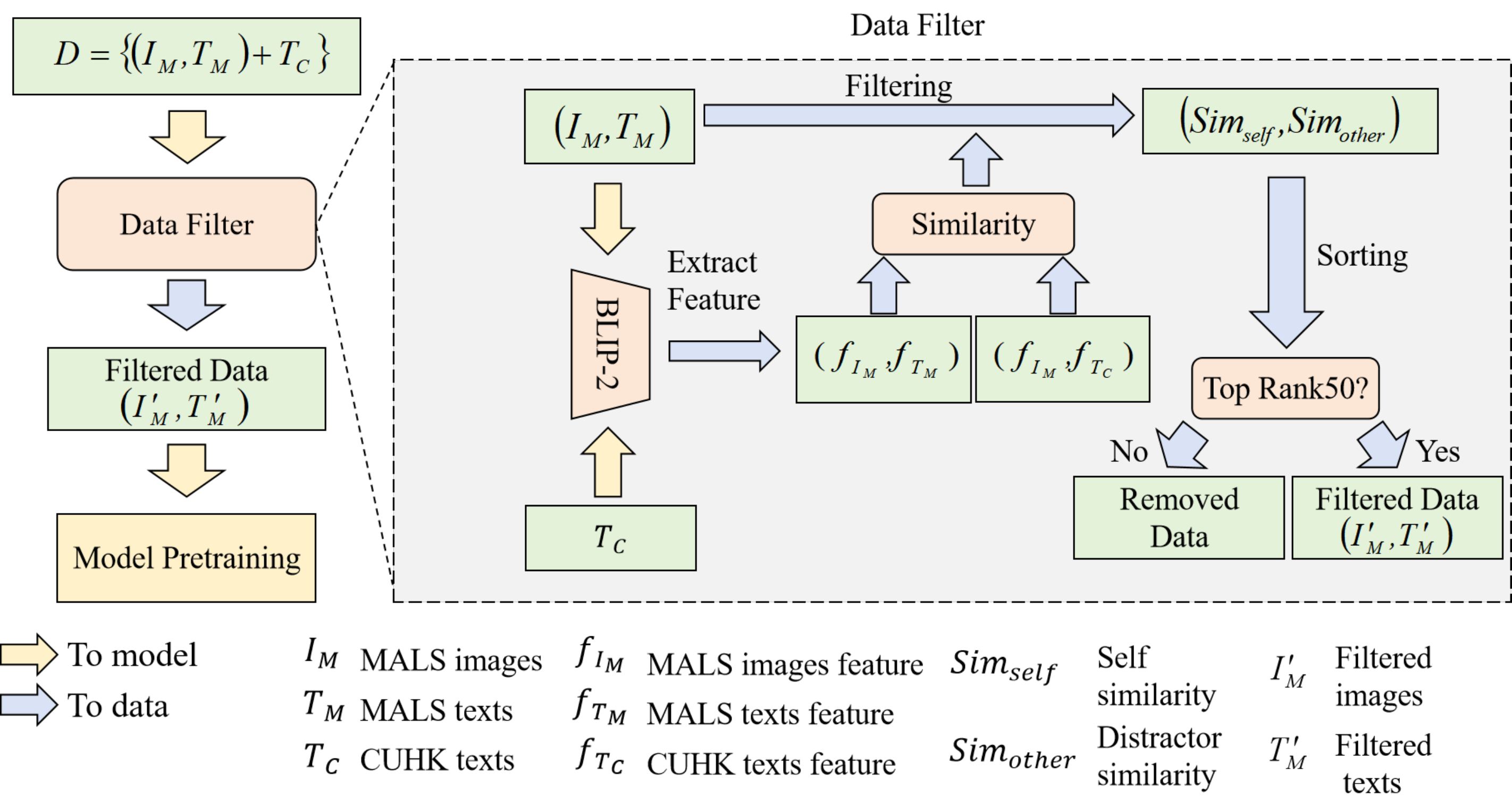}
    \vspace{-.3in}
  \caption{An overview of our data filtering process. We first employ Blip-2 \cite{li2023blip2} to extract features from the input image-text pair ${(I, T)}$ and the distractor text $T_{C}$. Next, we compute the similarity and rank the results accordingly, ultimately generating the filtered dataset. }
  \Description{An overview of data filter.}
\label{filter}
\vspace{-.15in}
\end{figure}

As highlighted in our introduction, the acquisition of images is challenged by high annotation costs and concerns over individual privacy and security, necessitating the generation of a large volume of image-text pairs ${(I, T)} $ through diffusion models, complemented by text descriptions generated by large language models~\cite{APTM}. Despite the potential for achieving high accuracy through extensive pretraining on such data, it becomes apparent that not all generated data are equally effective, with a significant portion being redundant. Additionally, these synthesized datasets often include noise in the form of image-text pairs with poor matching quality, which can inadequately represent the visual content of images. Such pairs serve as suboptimal signals for learning fine-grained visual-language alignment, potentially disrupting model training. This observation led us to ponder whether reducing the volume of data, focusing solely on a core set, could suffice for model training. Consequently, we devised a data filtering solution a data filtering approach based on the off-the-shelf large cross-modality model, \eg, BLIP-2~\cite{li2023blip2}, leveraging its efficient multimodal feature extraction capabilities to isolate features from both text and images (see Figure ~\ref{filter}). We intend to find the appropriate filtering methodology and thresholds to isolate a high-quality core set.

Our method is executed in two phases, starting with the filtering of the %MALS 
synthesized dataset, \ie, MALS, used for pretraining the model. Initially, all image-text pairs ${(I_{M}, T_{M})} $  within the dataset are processed. Given an image-text pair, the large cross-modality model, \ie, BLIP-2, is employed to extract the corresponding image feature $f_{I_{M}}$ and text feature $f_{T_{M}}$. Subsequently, the training set captions from the real-world downstream dataset, \ie, CUHK-PEDES, acting as distractor texts $T_{C}$, are utilized. For each pair, we calculate the cosine similarity between projected image embeddings and projected text embeddings to ascertain their self-similarity $Sim_{self}$. Simultaneously, the distractor similarity $Sim_{other}$ is determined by calculating the similarity between the image feature $f_{I_{M}}$ and a randomly selected subset of 10,000 distractor texts $T_{C}$. The following is the formula for calculating the similarity:
\setlength\abovedisplayskip{3pt}
\setlength\belowdisplayskip{3pt}
\begin{equation}
Sim_{self} =cos\left ( f_{I_{M} }, f_{T_{M}} \right ), 
Sim_{other} =cos\left ( f_{I_{M} }, f_{T_{C}} \right ),
\end{equation}
where $cos\left (,\right )$ denotes the cosine similarity. After computing these two types of similarities for all image-text pairs within the training dataset, and then we sort texts according to the similarity from highest to lowest, we set a ranking threshold as 50. It means that an image-text pair is retained only if its self-similarity ranks within the top 50 of all calculated similarities. 
Through this threshold-based filtering, we discarded 21\% of low-quality image-text pairs, retaining 79\% to form a new pretraining dataset, %Filter
Filtered-MALS.

Similarly, we also could filter the CUHK-PEDES training set in the fine-tuning phase to remove the noisy pair in the real-world training set. Employing the same calculation as in the first phase, we extract image features $f_ {I_ {C}} $ and text features $f_ {T_ {C}} $ for all pairs ${(I_{C}, T_{C})} $. Each image calculates the cosine similarity with both the ground-truth text and a subset of 10,000 random distractor texts $f_{T_{C'}}$. 
Similarly, the  similarity can be formulated as: %The following is the formula for calculating the similarity:
\setlength\abovedisplayskip{3pt}
\setlength\belowdisplayskip{3pt}
\begin{equation}
Sim_{self} =cos\left ( f_{I_{C} }, f_{T_{C}}\right ),
Sim_{other} =cos\left ( f_{I_{C} }, f_{T_{C'}}\right ).
\end{equation}
% After computing and ranking the similarities for the entire training set, we apply a relatively loose ranking threshold of 1800, considering that most pairs are human-annotated. Through this selection process, 10\% of low-quality image-text pairs are removed, retaining 90\% to form a new fine-tuning dataset, Refined-CUHK, for model fine-tuning.
Upon computing and ranking the similarities across the entire training set, we set a relatively loose ranking threshold of 1800, given that most pairs are human-annotated. This selection process results in the removal of 10\% low-quality image-text pairs, leaving 90\% to form a refined dataset, Refined-CUHK, for model fine-tuning.
% \begin{figure}[h]
%   \centering
%   \includegraphics[width=1.1\linewidth]{WoRA.png}
%   \caption{An overview of our proposed WoRA.The model is updated by fine-tuning the decomposition of the pre-trained weights into amplitude and direction components and updating both components using LoRA~\cite{hu2021lora} while adding the alpha and beta learnable parameters.}
%   \Description{A woman and a girl in white dresses sit in an open car.}
% \end{figure}
\begin{figure}[t]
\vspace{-.2in}
  \centering
  \includegraphics[width=\linewidth]{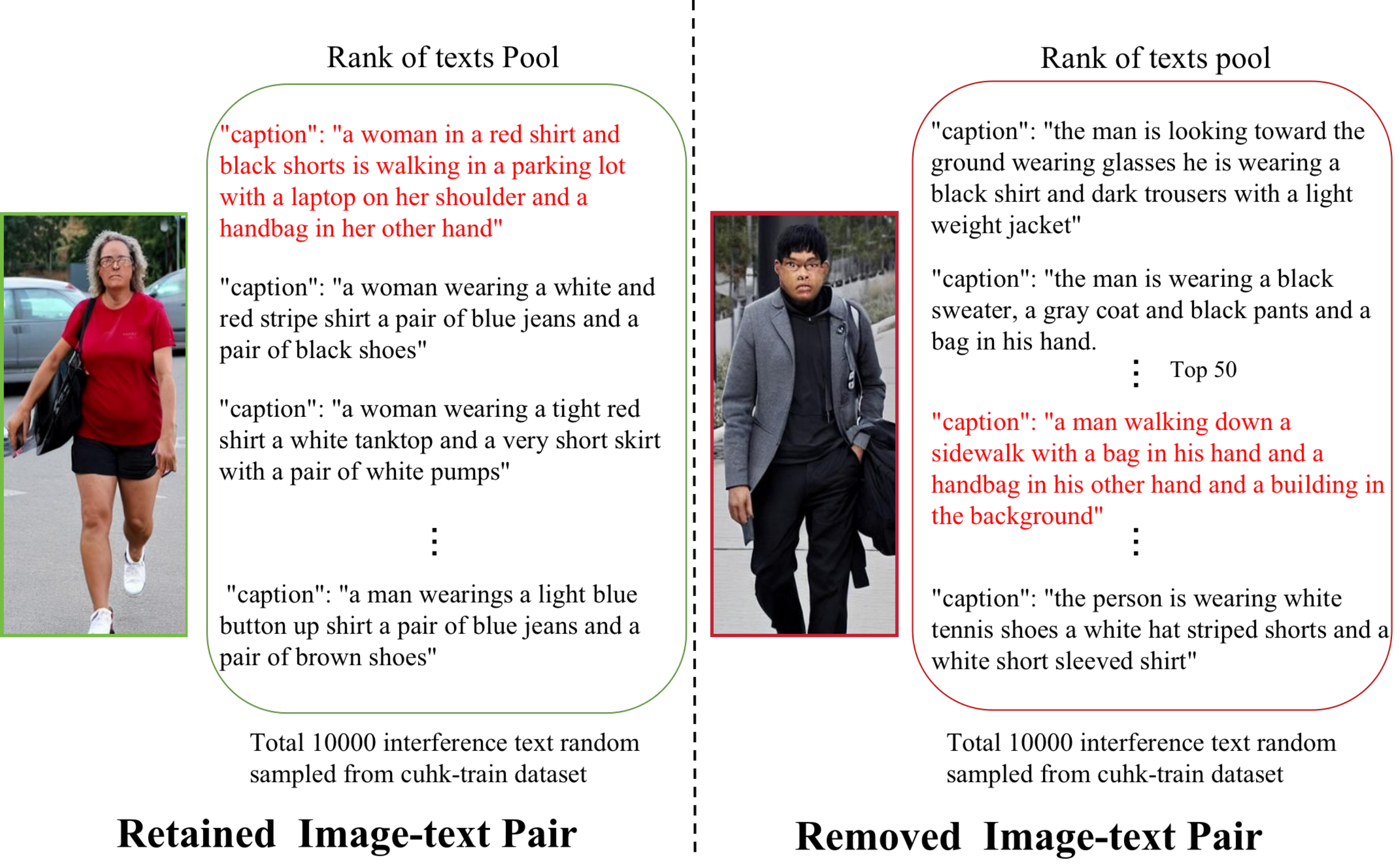}
    \vspace{-.3in}
  \caption{Visual explanation of data filtering. The part on the left of the image shows the high-quality image retained after our screening strategy and its corresponding red text description, while the person image on the right represents the low-quality image text pairs that are filtered out beyond the threshold, \ie, top50. We deploy the real-world training set as distractors to filter low-relevance synthesized image-text pairs according to the similarity since there are no overlaps.}
  \Description{Visual explanation of data filter.}
\label{Visual explanation of data filter}
\vspace{-1.2em}
\end{figure}

\vspace{-.1in}
\noindent\textbf{Discussion. The mechanism of filtering.} 
The motivation is within the spectrum of generated datasets. (1) It is worth noting that not all contents bear relevance, with a portion comprising low-quality image-text pairs. Such instances of subpar alignment between textual descriptions and corresponding images can invariably exert a deleterious impact, compromising the model training. Predominantly, the segments of the dataset that contribute most significantly to the performance of the model are those encompassing high-quality data, often referred to as the `coreset'. %Furthermore, 
(2) The extensive volume of data implicated in the pretraining phase incurs substantial computational costs, necessitating considerable resources in terms of computational power and temporal investment. Thus, data filtering provides two primary advantages: (1) Our filtering algorithm identifies the core set of image-text pairs that are most crucial for improving data quality during model training. This selective process ensures the model is exposed to higher-quality data, enhancing its learning capabilities. (2) Retaining only the core set significantly reduces the dataset size, thereby decreasing computational overhead during training. This results in reduced computational requirements and training time. Figure~\ref{Visual explanation of data filter} illustrates an example of the data filtering process.
% Therefore, there are two primary advantages of data filtering: (1) Initially, through the implementation of our filtering algorithm, we identify the core set of image-text pairs within the dataset that play a pivotal role in enhancing the quality of data fed into the model for learning. This strategic selection process facilitates the exposure of the model to higher-quality data, thereby augmenting its performance capabilities. (2) Subsequently, retaining the core set within the filtered dataset markedly reduces the overall volume of data, which, in turn, diminishes the computational overhead involved in the training process. This efficiency gain leads to a reduction in both the requisite computational power and the temporal expenditure. Figure ~\ref{Visual explanation of data filter} shows an example of the data filtering process.
% 生成数据集中的很多内容并不都是有效，其中可能包含很多低质量图像文本对，这些图文匹配质量不那么高的文本对往往会产生负面作用，使得模型学习性能下降，而往往数据集中真正有效的部分是我们称之为coreset的那部分高质量数据。同时大量的数据在预训练过程中会产生很高的计算成本，从而需要很多的算力和时间消耗。
% There are two primary advantages:
% （1）首先通过我们的过滤算法找到数据集中真正起关键作用的coreset 图像文本对会提升送入模型学习数据的质量，从而使得模型能够学到更多高质量的数据，提升模型性能。
% （2）其次过滤后的数据集保留coreset会大大减少数据的数量，从而减少训练过程中的计算成本，降低算力需求以及时间消耗。
\subsection{Weighted Low-Rank Adaptation}
%In pursuit of customizing generic models~\cite{zeng2022multigrained} for specific downstream tasks, full fine-tuning (FT) is commonly employed, entailing retraining of all model parameters. 
While the introduction of a pretrain-finetune paradigm to train models for person search tasks has achieved commendable results~\cite{APTM}, the expansion in model and dataset sizes significantly increases the number of parameters to be trained, demanding substantial computational resources and sacrificing training efficiency. To address this issue, several parameter-efficient fine-tuning methods~\cite{houlsby2019parameterefficient} have been proposed, aiming to fine-tune pretrained models using the minimum number of parameters. Among these, LoRA~\cite{hu2021lora} has gained popularity due to its simplicity and efficacy. LoRA employs a low-rank decomposition for the pretrained weight matrix $W_{0} \in \mathbb{R} ^{m\times n} $, $W_{0} + \bigtriangleup W = W_{0} + BA $, where $B \in \mathbb{R} ^{m\times r} $, $A \in \mathbb{R} ^{r\times n}$, rank $r \ll \min_{}\left ( m,n \right )$. $\bigtriangleup W $ is adjusted by $\frac{8}{r} $. Inspired by the LoRA~\cite{hu2021lora}, which updates only a small part of the model weight to improve efficiency, DoRA~\cite{liu2024dora} decomposes the weight into two parts: direction and amplitude. DoRA improves the adaptability and efficiency of the model, which can be formulated as $W_{DoRA}  = m\frac{W_{0}+ BA  }{\left \| W_{0}+ BA \right \| _{2} },$ where $m \in \mathbb{R} ^{1\times n} $ is the magnitude vector. $ \left \| \cdot  \right \| _{2} $ denotes the L2 norm of a matrix across each column.
%Inspired by this decompose and merge manner, 

However, these two methods still limit the parameter freedom. Therefore, we introduce the Weighted Low-Rank Adaptation (WoRA) model to address the capacity gap still present between LoRA~\cite{hu2021lora} and fine-tuning (see Figure~\ref{arte} right). % often attributed to the limited number of trainable parameters. 
Drawing from the DoRA~\cite{liu2024dora} approach, which reparameterizes model weights into magnitude and direction components for fine-tuning, WoRA introduces new learnable parameters, $\alpha$ and $\beta$, to facilitate parameter learning. Given that pretrained weights already possess a vast repository of knowledge suitable for various downstream tasks, we configure these learnable parameters to ensure the model acquires sufficient capability in both magnitude and direction. This allows for the model to adapt to downstream tasks by updating parameters that exhibit significant magnitude or directional changes. The formal representation of our WoRA model is:
\begin{equation}
W_{WoRA}  = m \frac{ \beta  * W_{0}+ \alpha  * BA}{\left \| \beta  *W_{0}+ \alpha *BA \right \|_{2} }. 
\end{equation}
In our model, parameters denoted with an overline represent trainable parameters. The proposed Weighted Low-Rank Adaptation (WoRA) demonstrates learning capabilities comparable to full fine-tuning. During inference, WoRA integrates with pretrained weights, introducing no additional latency while enhancing both learnability and computational efficiency. Moreover, by incorporating new learnable parameters, WoRA achieves superior performance compared to LoRA~\cite{hu2021lora} and DoRA~\cite{liu2024dora}.
% In our model, parameters denoted with an overline represent trainable parameters. Our Weighted Low-Rank Adaptation (WoRA) achieves learning capabilities remarkably similar to those of fine-tuning. During inference, WoRA merges with pretrained weights, thereby not introducing any additional latency and enhancing the learnability and computational speed of the model. Furthermore, through the introduction of new learnable parameters, our WoRA indicates superior learning performance compared to both LoRA~\cite{hu2021lora} and DoRA~\cite{liu2024dora}.
% \begin{figure}[t]
%   \centering
%   \includegraphics[width=\linewidth]{fig5new.png}
%   \caption{Intuitive explanation of WoRA. (a) We show weight changes for LoRA and DoRA (b) We show weight changes for WoRA, where $BA$ represents weight direction rotation, and $\alpha, \beta,$ and $m$ take charge of magnitude.} %\zznote{Please make WoRA different from DoRA, such as smaller beta, bigger alpha. Re-draw figure.} }
%   \Description{Visual explanation of WoRA.}
% \label{Visual explanation of WoRA}
% \vspace{-0.5em}
% \end{figure}
\begin{figure}[t]
  \centering
  \vspace{-.4in}
  \includegraphics[width=\linewidth]{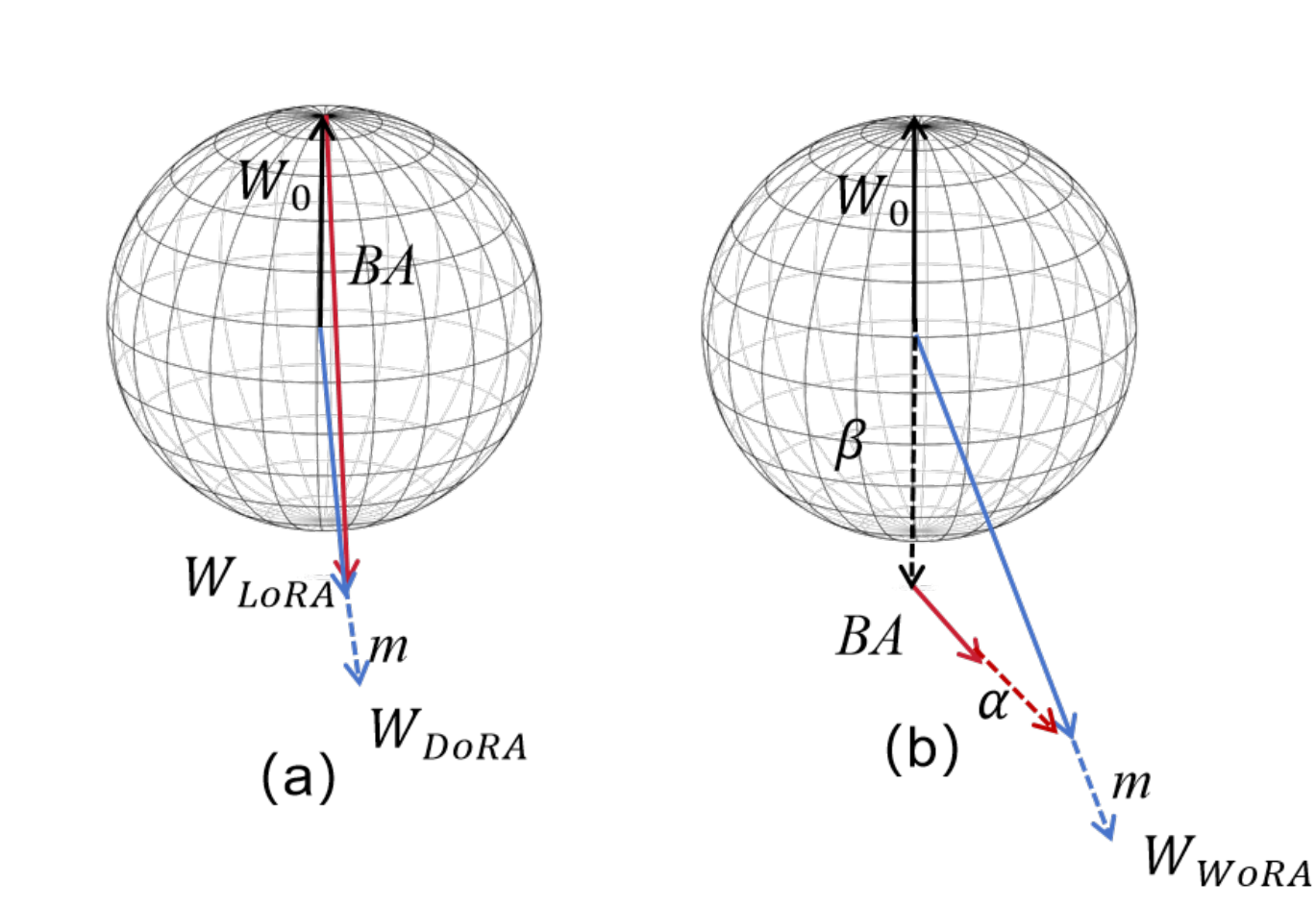}
  \vspace{-.35in}
  \caption{Intuitive comparison of LoRA, DoRA, and our proposed WoRA. (a) Here we show a common case during optimization, \ie, negative correlation against $W_{0}$, which both LoRA and DoRA are struggling with. The bias parameter $BA$ is hard to learn, considering the weight decay and other regularization. 
  (b) In contrast, we deploy two float scalars in WoRA, \ie, $\alpha$ and $\beta$, which could efficiently adjust the vector and provide better flexibility.} %\zznote{Please make WoRA different from DoRA, such as smaller beta, bigger alpha. Re-draw figure.} }
  \Description{Visual explanation of WoRA.}
\label{Visual explanation of WoRA}
\vspace{-1.5em}
\end{figure}

\vspace{1mm}
\noindent\textbf{Discussion. Why is WoRA better than LoRA and DoRA?} 
% Both DoRA and LoRA are the specific cases of our proposed WoRA. 
% In particular, DoRA can be derived by WoRA via setting $\alpha = 1$ and $\beta = 1$ as fixed constants. 
% Further, if we set WoRA $\alpha = 1$, $\beta = 1$ and $m = 1$ as fixed constants, we will achieve the identical function as LoRA. 
% Let us think of a negative correlation case, which is also common during training. As shown in Figure ~\ref{Visual explanation of WoRA}~(a), we can see that both LoRA and DoRA are struggling with the common negative correlation against the original weight $W_0$. They need to learn a large $BA$, but it is usually challenging and unstable, considering weight decay and other regularization terms. 
% In contrast, by introducing two float scalars, \ie, $\alpha$ $\beta$, we can achieve rapid adaptation and a more flexible space for change. 
% Our method can easily achieve negative correlation by setting a negative $\beta$ see Figure ~\ref{Visual explanation of WoRA}~(b). 
% Similarly, $\alpha$ could control the fine control over $BA$ during optimization. We will add the illustration to the revision.
Both DoRA and LoRA are specific cases of the proposed WoRA. Specifically, DoRA can be derived from WoRA by setting $\alpha = 8$ and $\beta = 1$ as fixed constants. Similarly, setting $\alpha = 8$, $\beta = 1$, and $m = 1$ yields the identical behavior as LoRA.
Consider a scenario involving negative correlation, which is common during training. As illustrated in Figure~\ref{Visual explanation of WoRA}~(a), both LoRA and DoRA struggle with the typical negative correlation against the original weight $W_0$, necessitating \textbf{the learning of a large $BA$—a challenging and unstable process due to weight decay and other regularization terms}.
In contrast, by introducing two scalar parameters, $\alpha$ and $\beta$, WoRA provides a more flexible and adaptive space for optimization. Our method can readily achieve negative correlation by setting a negative $\beta$, as depicted in Figure~\ref{Visual explanation of WoRA}~(b). Additionally, $\alpha$ offers fine-grained control over $BA$ during optimization. We will add an illustration of this to the revised version.
\textbf{Time Cost}: WoRA introduces three additional trainable parameters compared to LoRA, resulting in a slightly longer training time due to the increased complexity of weight calculation. Although WoRA's training time is marginally higher than that of DoRA and LoRA, the additional trainable parameters enable better weight adjustment, ultimately enhancing performance. As shown in Table~\ref{tab:method}, a small increase in training time (approximately 30 minutes) significantly improves the model's fine-tuning capability.
\textbf{Memory Cost}: It is noteworthy that our space complexity is equivalent to DoRA, while WoRA introduces greater freedom in direction and magnitude, thereby facilitating fine-tuning. WoRA adds only two constant learnable scalars per weight update compared to DoRA, resulting in a negligible increase in space complexity. 
\section{Experiment}
\subsection{Experimental Setup}
% \textbf{Datasets}. Our study encompasses processing and training on four datasets. For pretraining, we utilize the synthetic dataset, \ie, MALS~\cite{APTM}. The MALS dataset contains 1,510,330 image-text pairs, with each pair annotated with appropriate attribute labels. For the fine-tuning phase, we employ CUHK-PEDES~\cite{Li_Xiao_Li_Zhou_Yue_Wang_2017}, RSTPReid~\cite{zhu2021dssl} and ICFG-PEDES. CUHK-PEDES aggregates 40,206 images of 13,003 individuals from five existing person search datasets: CUHK03~\cite{Li_Zhao_Xiao_Wang_2014}, Market-1501~\cite{Zheng_Shen_Tian_Wang_Bu_Tian_2015}, SSM~\cite{Xiao_Li_Wang_Li_Wang_2016}, VIPER~\cite{Gray_Brennan_Tao_2007}, and CUHK01~\cite{Li_Zhao_Wang_2013}, serving as subjects for language descriptions. Each image is annotated with descriptions from two sentences, amassing a total of 80,412 sentences. Our approach is evaluated on the public text-based person search dataset CUHK-PEDES. RSTPReid comprises 20,505 images of 4,101 people and is created by compiling MSMT17~\cite{Wei_Zhang_Gao_Tian_2018}. ICFG-PEDES~\cite{Ding_Ding_Shao_Tao_2021} is also created from MSMT17 and has 54,522 images of 4,102 individuals.
\textbf{Datasets}. Our study involves processing and training on four datasets. For pretraining, we utilize the synthetic dataset, MALS~\cite{APTM}, which comprises 1,510,330 image-text pairs, each annotated with relevant attribute labels. For fine-tuning, we employ CUHK-PEDES~\cite{Li_Xiao_Li_Zhou_Yue_Wang_2017}, RSTPReid~\cite{zhu2021dssl}, and ICFG-PEDES. CUHK-PEDES aggregates 40,206 images of 13,003 individuals from 5 existing person reID datasets: CUHK03~\cite{Li_Zhao_Xiao_Wang_2014}, Market-1501~\cite{Zheng_Shen_Tian_Wang_Bu_Tian_2015}, SSM~\cite{Xiao_Li_Wang_Li_Wang_2016}, VIPER~\cite{Gray_Brennan_Tao_2007}, and CUHK01~\cite{Li_Zhao_Wang_2013}. Each image is paired with descriptions from two sentences, resulting in a total of 80,412 sentences. Our approach is evaluated on the public text-based person search dataset CUHK-PEDES. RSTPReid contains 20,505 images of 4,101 individuals, compiled from MSMT17~\cite{Wei_Zhang_Gao_Tian_2018}. ICFG-PEDES~\cite{Ding_Ding_Shao_Tao_2021}, also derived from MSMT17, consists of 54,522 images of 4,102 individuals.

% \vspace{1mm}
\noindent\textbf{Evaluation metrics}. We adopt the mean Average Precision (AP) and Recall@1,5,10 as our primary evaluation metrics. The Recall@K, whose value is 1 if the first matched image has appeared before the K-th image. Recall@K is sensitive to the position of the first matched image and suits the test set with only one true-matched image in the gallery. The average precision (AP) is the area under the PR(Precision-Recall) curve, considering all ground-truth images in the gallery. mAP is calculated and averaged for the average accuracy (AP) of each category.
%Accuracy refers to the proportion of the actual positive samples to the predicted positive samples among the samples predicted by the model. mAP is calculated and averaged for the average accuracy (AP) of each category. The higher the mAP value, the better the performance of the model.

\noindent\textbf{Implementation Details}. Our model with the proposed WoRA, undergoes pretraining across 32 epochs on 8 NVIDIA A800 GPUs via Pytorch, adopting a batch size of 150. The optimization strategy employs the AdamW optimizer~\cite{Loshchilov_Hutter_2017}, integrating a weight decay factor of 0.01. Initiation of the learning rate is set at $1e^{-5}$, incorporating a warm-up phase over the initial 2600 steps, which then transitions into a linear decay schedule ranging from $1e^{-4}$ down to $1e^{-5}$. Image preprocessing includes resizing to 384 × 128 dimensions, coupled with augmentation strategies such as random horizontal flipping, RandAugment~\cite{Cubuk_Zoph_Shlens_Le_2020}, and random erasing~\cite{Zhong_Zheng_Kang_Li_Yang_2020}. For the pretraining phase, the image encoder is initialized using Swin-B~\cite{SwinTransformer}, enhanced by the application of the WoRA method. Both the text encoder and cross encoder commence with configurations derived from the initial and final 6 layers of Bert~\cite{devlin2019bert}, respectively. Subsequent to the pretraining, the model is fine-tuned on designated downstream datasets over 30 epochs. Initial WoRA settings for the image encoder, rank = 8, $\alpha$ = 8, and $\beta$ = 1, are preserved, with the learning rate commencing at $1e^{-4}$. This phase includes a warm-up period spanning the first three epochs, succeeded by a methodical reduction of the learning rate according to a linear scheduler. In addition to image data augmentation mentioned during pretraining, Easy Data Augmentation (EDA)~\cite{Wei_Zou_2019} is employed for text data augmentation, with the batch size as 120. For each text query, its cosine similarity with all images is calculated, selecting the top 128 candidate images. Subsequently, the match probability between the text query and each selected image candidate is computed and ranked~\cite{APTM}.
\begin{table}[t]
\vspace{-.15in}
\fontsize{8}{7}\selectfont
\setlength{\tabcolsep}{5pt}
\renewcommand{\arraystretch}{1.0}
\scalebox{1.0}{\small
\begin{tabular}{lccccc}
\toprule
\bf Method    & \bf \#Parameter       & \bf R@1   & \bf R@5   & \bf R@10  & \bf mAP   \\ \hline
CNN-RNN~\cite{reed2016learning}                           & -   & 8.07  & -     & 32.47 & -                \\
GNA-RNN~\cite{Li_Xiao_Li_Zhou_Yue_Wang_2017}         & -  & 19.05 & -     & 53.64 & -                \\
PWM-ATH~\cite{8354312}                                     & -  & 27.14 & 49.45 & 61.02 & -                \\
GLA~\cite{chen2018improving}                               & -  & 43.58 & 66.93 & 76.2  & -               \\
Dual Path~\cite{Zheng_Zheng_Garrett_Yang_Xu_Shen_2020} & - & 44.40 & 66.26 & 75.07 & -             \\
CMPM+CMPC~\cite{10.1007/978-3-030-01246-5_42}            & -   & 49.37 & -     & 79.21 & -               \\
MIA~\cite{niu2019improving}                               & -   & 53.10 & 75.00 & 82.90 & -                \\
A-GANet~\cite{10.1145/3343031.3350991}                    & -   & 53.14 & 74.03 & 81.95 & -               \\
ViTAA~\cite{wang2020vitaa}                                & 177M   & 55.97 & 75.84 & 83.52 & 51.60            \\
IMG-Net~\cite{Wang2020IMGNetIA}                           & -   & 56.48 & 76.89 & 85.01 & -                \\
CMAAM~\cite{9093640}                                      & -   & 56.68 & 77.18 & 84.86 & -               \\
HGAN~\cite{10.1145/3394171.3413864}                       & -   & 59.00 & 79.49 & 86.62 & -               \\
NAFS~\cite{gao2021contextual}                             & 189M   & 59.94 & 79.86 & 86.70 & 54.07            \\
DSSL~\cite{zhu2021dssl}                                   & -   & 59.98 & 80.41 & 87.56 & -              \\
MGEL~\cite{Wang2021TextbasedPS}                           & -   & 60.27 & 80.01 & 86.74 & -                \\
SSAN~\cite{Ding_Ding_Shao_Tao_2021}                   & -   & 61.37 & 80.15 & 86.73 & -              \\
NAFS~\cite{gao2021contextual}                          & 189M      & 61.50 & 81.19 & 87.51 & -                \\
TBPS~\cite{han2021textbased}                            & 43M     & 61.65 & 80.98 & 86.78 & -               \\
TIPCB~\cite{Chen_Zhang_Lu_Wang_Zheng_2022}           & 185M   & 63.63 & 82.82 & 89.01 & -               \\
LBUL~\cite{Wang_Zhu_Xue_Wan_Liu_Wang_Li_2022}       & -  & 64.04 & 82.66 & 87.22 & -               \\
CAIBC~\cite{CAIBC}                                       & -    & 64.43 & 82.87 & 88.37 & -               \\
AXM-Net~\cite{farooq2022axmnet}                          & -    & 64.44 & 80.52 & 86.77 & 58.73           \\
LGUR~\cite{Shao_Zhang_Fang_Lin_Wang_Ding_2022}     & -    & 65.25 & 83.12 & 89.00 & -               \\
CFine~\cite{yan2022clipdriven}                          & -     & 69.57 & 85.93 & 91.15 & -                \\
VGSG~\cite{10345496}                                     & -    & 71.38 & 86.75 & 91.86 & 67.91            \\
TBPS-CLIP~\cite{cao2023empirical}                        & 149M    & 73.54 & 88.19 & 92.35 & 65.38         \\
IRRA~\cite{jiang2023crossmodal}                          & 194M    & 73.38 & 89.93 & 93.71 & 66.13        \\
RaSa~\cite{Bai_2023}                                     & 210M   & 76.51 & 90.29 & 94.25 & 69.38        \\
APTM~\cite{APTM}                                         & 214M     & 76.53 & 90.04 & 94.15 & 66.91       \\ \hline
Baseline*                                                  & 214M                   & 75.42 & 88.86 & 92.77 & 66.61        \\
\bf Ours                                                       & \bf 127M                   & \bf 76.38 & \bf 89.72 & \bf 93.49 & \bf 67.22         \\ \bottomrule
\end{tabular}
}
\caption{Performance Comparison on CUHK-PEDES. Here we show the performance of the previous methods on the recall@1,5,10, mAP in \% and the parameter number. Baseline*: We re-implement APTM~\cite{APTM}.  }
\label{tab:total}
\vspace{-.40in}
\end{table}
\vspace{-.1in}
\subsection{Comparison with Competitive Methods}
\begin{figure}[t]
\vspace{-.1in}
  \centering
  \includegraphics[width=\linewidth]{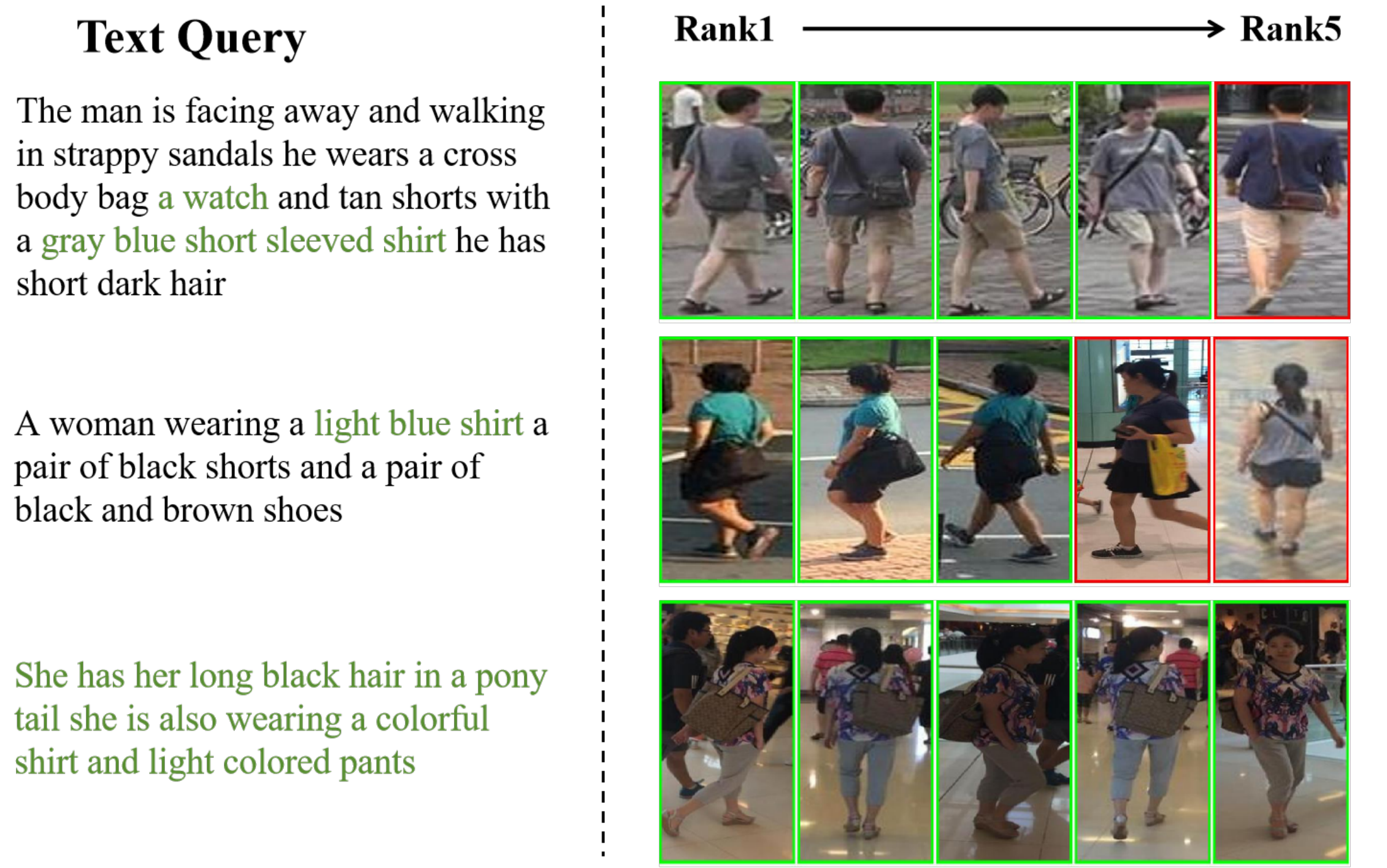}
  \vspace{-.30in}
  \caption{Qualitative person search results using text query of our methods, placing in descending order from left to right based on matching probability. %The results are ranked from left to right by similarity score, from r1 to r5. 
  The images in \textcolor{green}{green} boxes are the correct matches, and the images in \textcolor{red}{red} boxes are the wrong matches. The green texts show that our results successfully match. }
  \Description{Qualitative result.}
\label{qualitative}
\vspace{-1.0em}
\end{figure}

We deploy the Weighted Low-Rank Adaptation (WoRA) for text-based person search tasks. Performance comparisons are made using Recall@1,5,10 and mean Average Precision (mAP) metrics, alongside a comparison of parameter count (params in Millions, M) and computational efficiency (FLOPs) against the baseline model APTM. 
%WoRA is evaluated on the CUHK-PEDES, RSTPReid and ICFG-PEDES datasets. 
Through trainable adjustments to weight parameters, the WoRA method indicated robust performance across both datasets, significantly reducing computational parameters and time. Specifically, compare to APTM, which is trained on 1.51M data, our data filtering algorithm remove 21\% of low-quality, noisy data, utilizing 1.19M data for computations. Our implementation of WoRA on the CUHK-PEDES dataset reduce trainable parameters to 127.37M, a 41\% decrease from APTM, with FLOPs reduce to 23.21G, a 39\% reduction. The overall training duration for pretraining and fine-tuning is cut by 19.82\%, with our model achieving slight improvements in recall rates and mAP, as evidenced in Table ~\ref{tab:total}. Moreover, the pretrained model adjusted through WoRA achieves competitive performance on the RSTPReid and ICFG-PEDES dataset, as shown in Table ~\ref{tab:RSTP} and ~\ref{tab:icfg}. To show the performance of our model more intuitively, we present three visual retrieval results on CUHK-PEDES in Figure ~\ref{qualitative}. Our model adeptly captures fine-grained, word-level details, enabling it to accurately differentiate subtle variations in clothing colors among individuals. Furthermore, it exhibits robust retrieval capabilities, effectively identifying subjects even when parts of their details are obscured. This indicates the strong performance of our model in handling nuanced visual variations and partial occlusions within complex scenes.

\begin{table}[t]
\vspace{-.1in}
\fontsize{8}{8}\selectfont
\setlength{\tabcolsep}{6pt}
\renewcommand{\arraystretch}{1.0}
\scalebox{1.0}{\small
\begin{tabular}{lccccc}
\toprule
\bf Method  & \bf \#Parameter & \bf R@1   & \bf R@5   & \bf R@10  & \bf mAP    \\ \hline
DSSL~\cite{zhu2021dssl}  & -  & 32.43 & 55.08 & 63.19 & -      \\
LBUL~\cite{Wang_Zhu_Xue_Wan_Liu_Wang_Li_2022} & -  & 45.55 & 68.20 & 77.85 & -      \\
IVT~\cite{Shu_Wen_Wu_Chen_Song_Qiao_Ren_Wang_2022}  & -  & 46.70 & 70.00 & 78.80 & -      \\
CAIBC~\cite{CAIBC} & - & 47.35 & 69.55 & 79.00 & -      \\
CFine~\cite{yan2022clipdriven} & - & 50.55 & 72.50 & 81.60 & -      \\
TBPS-CLIP~\cite{cao2023empirical} & 149M &61.95	&83.55	&88.75	&48.26  \\
IRRA~\cite{jiang2023crossmodal} & 194M & 60.20 & 81.30 & 88.20 &47.17      \\
RaSa~\cite{Bai_2023} & 210M  & 66.90 & 86.50 & 91.35 & 52.31  \\
APTM~\cite{APTM}  & 214M  & 67.50 & 85.70 & 91.45 & 52.56  \\
\hline
Baseline*   & 214M   & 66.40 & 85.55 & 91.10 & 52.21   \\                   
\bf Ours  & \bf 127M & \bf 66.85 & \bf 86.45 & \bf 91.10 & \bf 52.49  \\ 
\bottomrule
\end{tabular}
}
\caption{Performance Comparison on RSTPReid. Here we show the performance of the previous methods on the recall@1,5,10 and mAP in \% and the parameter number. Baseline*: We re-implement APTM~\cite{APTM}.  }
\label{tab:RSTP}
\vspace{-.35in}
\end{table}
\begin{table}[t]
% \vspace{-.05in}
\fontsize{8}{8}\selectfont
\setlength{\tabcolsep}{5pt}
\renewcommand{\arraystretch}{1.0}
\small
\begin{tabular}{lccccc}
\toprule
\bf Method  & \bf \#Parameter  & \bf R@1    & \bf R@5    & \bf R@10   & \bf mAP    \\ \hline
Dual Path~\cite{Zheng_Zheng_Garrett_Yang_Xu_Shen_2020} & - & 38.99  & 59.44  & 68.41  & -     \\
CMPM+CMPC~\cite{10.1007/978-3-030-01246-5_42}  & - & 43.51  & 65.44  & 74.26  & -    \\
MIA~\cite{niu2019improving}    & -   & 46.49  & 67.14  & 75.18  & -     \\
SCAN~\cite{lee2018stacked}   & -  & 50.05  & 69.65  & 77.21  & -     \\
ViTAA~\cite{wang2020vitaa}  & 177M  & 50.98  & 68.79  & 75.78  & -     \\
SSAN~\cite{Ding_Ding_Shao_Tao_2021}    & - & 54.23  & 72.63  & 79.53  & -      \\
IVT~\cite{Shu_Wen_Wu_Chen_Song_Qiao_Ren_Wang_2022}    & -  & 56.04  & 73.60  & 80.22  & -      \\
LGUR~\cite{Shao_Zhang_Fang_Lin_Wang_Ding_2022}   & -  & 59.02  & 75.32  & 81.56  & -      \\
CFine~\cite{yan2022clipdriven}   & -  & 60.83  & 76.55  & 82.42  & -      \\
TBPS-CLIP~\cite{cao2023empirical} & 149M &65.05  &80.34	&85.47	&39.83  \\
IRRA~\cite{jiang2023crossmodal}   & 194M  & 63.46  & 80.25  & 85.82  & 38.06   \\
RaSa~\cite{Bai_2023}   & 210M  & 65.28  & 80.04  & 85.12  & 41.29  \\
APTM~\cite{APTM} & 214M  & 68.51  & 82.99  & 87.56  & 41.22  \\
\hline
Baseline*   & 214M & 67.81  & 82.70  & 87.32  & 41.22  \\
\bf Ours    & \bf 127M   & \bf 68.35 & \bf 83.10 & \bf 87.53 & \bf 42.60  \\ 
\bottomrule
\end{tabular}
\caption{Performance Comparison on ICFG-PEDES. Here we show the performance of the previous methods on the recall@1,5,10 and mAP in \% and the parameter number. Baseline*: We re-implement APTM~\cite{APTM}.  }
\label{tab:icfg}
\vspace{-.35in}
\end{table}

\begin{table*}[t]
\vspace{-.15in}
\fontsize{8}{8}\selectfont
    \setlength{\tabcolsep}{6pt}
    \renewcommand{\arraystretch}{1.0}
\begin{tabular}{ccccccccccc}
\toprule
\multicolumn{1}{c|}{Method}                       & \multicolumn{1}{c|}{\# Data (M)}        & \multicolumn{1}{c|}{\# Trainable}         & \multicolumn{1}{c|}{Flops (G)}        & \multicolumn{2}{c|}{CUHK-PEDES}                      & \multicolumn{2}{c|}{RSTPReid}                       & \multicolumn{2}{c|}{ICFG-PEDES}                       & \multicolumn{1}{c}{Time}   \\
\multicolumn{1}{c|}{}                             & \multicolumn{1}{c|}{}                & \multicolumn{1}{c|}{Params (M)}       & \multicolumn{1}{c|}{}                 & R@1            & \multicolumn{1}{c|}{mAP}            & R@1            & \multicolumn{1}{c|}{mAP}           & R@1             & \multicolumn{1}{c|}{mAP}            & \multicolumn{1}{c}{(hours)}  \\ \midrule
Pretraining Stage                                 &                                      &                                        &                                       &                &                                     &                &                                    &                 &                                     &              \\ \midrule
\multicolumn{1}{c|}{Baseline* (APTM~\cite{APTM})} & \multicolumn{1}{c|}{1.51M}           & \multicolumn{1}{c|}{213.99 M}          & \multicolumn{1}{c|}{38.02 G}          & 3.99           & \multicolumn{1}{c|}{3.62}           & 4.40           & \multicolumn{1}{c|}{3.95}          & 0.77            & \multicolumn{1}{c|}{0.59}           & 18h          \\
\multicolumn{1}{c|}{Ours (top50)}                 & \multicolumn{1}{c|}{1.19M}           & \multicolumn{1}{c|}{213.99 M}          & \multicolumn{1}{c|}{38.00 G}          & 10.25          & \multicolumn{1}{c|}{10.24}          & 12.65          & \multicolumn{1}{c|}{9.37}          & 8.07           & \multicolumn{1}{c|}{2.35}          & 14h          \\
\multicolumn{1}{c|}{Ours (top50+WoRA)}            & \multicolumn{1}{c|}{\textbf{1.19M}}  & \multicolumn{1}{c|}{\textbf{127.37 M}} & \multicolumn{1}{c|}{\textbf{23.21 G}} & \textbf{10.71} & \multicolumn{1}{c|}{\textbf{10.33}} & \textbf{13.00} & \multicolumn{1}{c|}{\textbf{9.59}} & \textbf{10.80} & \multicolumn{1}{c|}{\textbf{3.10}} & 14h*          \\ \midrule
Finetune Stage                                    &                                      &                                        &                                       &                &                                     &                &                                    &                 &                                     &              \\ \midrule
\multicolumn{1}{c|}{Baseline* (APTM~\cite{APTM})} & \multicolumn{1}{c|}{0.068M}          & \multicolumn{1}{c|}{213.99 M}          & \multicolumn{1}{c|}{44.93 G}          & 75.42          & \multicolumn{1}{c|}{66.61}          & 66.32          & \multicolumn{1}{c|}{52.30}         & 67.66           & \multicolumn{1}{c|}{41.98}          & 4.2h      \\
\multicolumn{1}{c|}{Ours (top50)}                & \multicolumn{1}{c|}{0.061M}          & \multicolumn{1}{c|}{213.99 M}          & \multicolumn{1}{c|}{44.93 G}          & 75.67          & \multicolumn{1}{c|}{66.27}          & 66.40          & \multicolumn{1}{c|}{52.21}         & 67.80           & \multicolumn{1}{c|}{42.38}          & 3.8h      \\
\multicolumn{1}{c|}{Ours (top50+WoRA)}            & \multicolumn{1}{c|}{\textbf{0.061M}} & \multicolumn{1}{c|}{\textbf{127.37 M}} & \multicolumn{1}{c|}{\textbf{30.13 G}} & \textbf{76.38}          & \multicolumn{1}{c|}{\textbf{67.22}} & \textbf{66.85}          & \multicolumn{1}{c|}{\textbf{52.49}}         & \textbf{68.35}           & \multicolumn{1}{c|}{\textbf{42.60}} & 3.8h*      \\ 
\bottomrule
\end{tabular}
\caption{Compared with APTM method at recall@1 and mAP results on CUHK-PEDES, RSTPReid and ICFG-PEDES. Meanwhile, we also compare the data volume, params (M) and Flops (G) of the model. Ablation study about our methods on pretrain. The top50 denotes the results trained by using a pre-trained dataset filtered using a data filtering method. The top50+WoRA is our ultimate two-stage approach, and by adding WoRA to fine-tune the model, we can improve the performance of the pre-trained model while saving 19.82\% training time. Time* indicates that the time decrease significantly because the bottleneck of model at other places. Thus, even though we optimize the GPU FPS, it have minimal impact on the overall computation time.}
\label{tab:pretrain}
\vspace{-2.3em}
\end{table*}
\begin{figure}[t]
\vspace{-.05in}
  \centering
  \includegraphics[width=\linewidth]{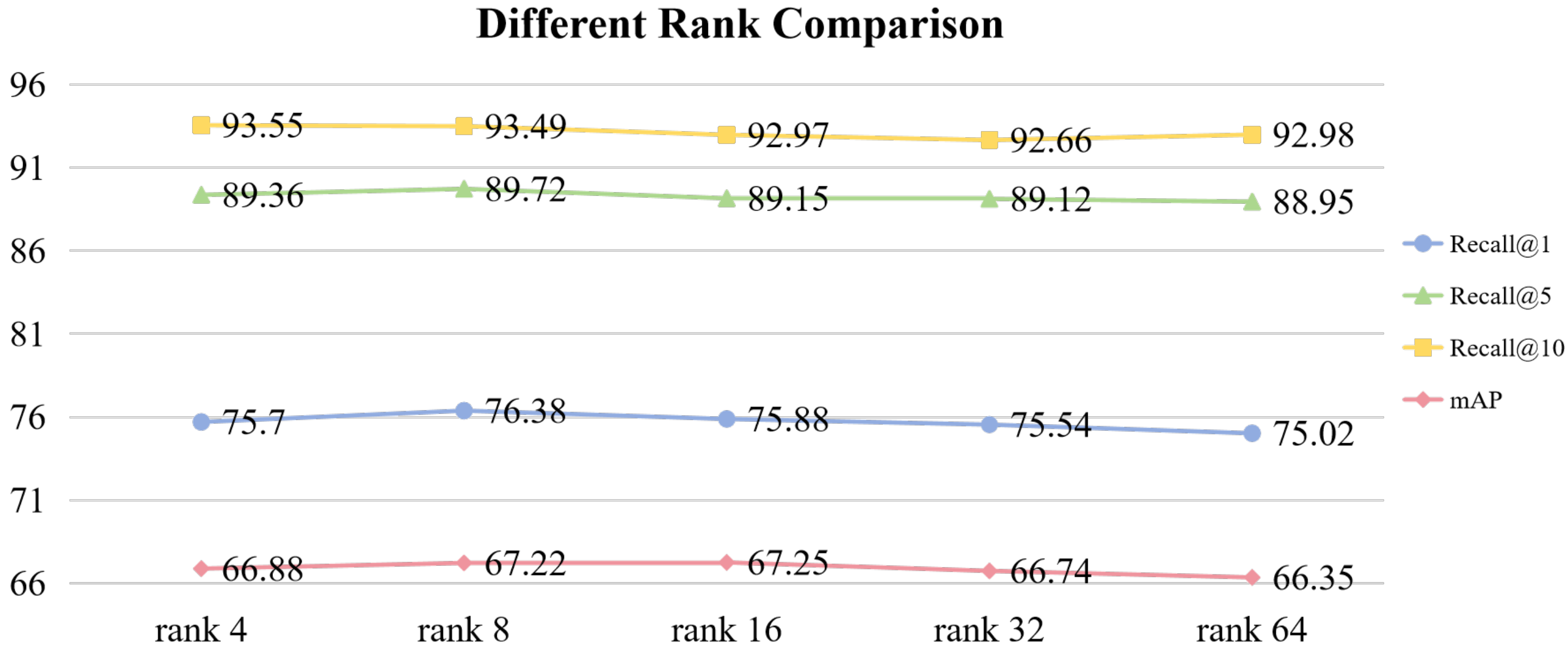}\vspace{-.1in}
  \caption{The impact of different WoRA ranks on performance. We observe that the result is not sensitive to the rank. Generally, rank=8 is the best hyper-parameter in terms of performance on Recall@1,5,10.}
  \Description{The impact of different ranks.}
  
\label{rank}
\vspace{-.20in}
\end{figure}
% \begin{table}[t]
%     \setlength{\tabcolsep}{10pt}
%     \renewcommand{\arraystretch}{1.3}
% \scalebox{1.0}{\small
% \begin{tabular}{c|cccc}
% \toprule
% Method   & top50 & ft 90\% & R@1    & mAP    \\ \hline
% Baseline* &       &        & 75.42 & 66.61 \\
% Baseline* & \Checkmark    &        & 75.83 & 66.70 \\
% Baseline* & \Checkmark     & \Checkmark      & 75.67 & 66.27 \\
% LoRA     & \Checkmark     &        & 74.40 & 64.95 \\
% LoRA     & \Checkmark     & \Checkmark      & 74.29 & 65.59 \\
% DoRA     & \Checkmark     & \Checkmark      & 75.73 & 66.75 \\ \hline
% WoRA (Ours)     & \Checkmark     & \Checkmark      & 76.38 & 67.22 \\ 
% \bottomrule
% \end{tabular}
% }
% \caption{Comparison of our WoRA with baseline, LoRA and DoRA in the different situations. Finally, the experiment shows that our methods achieve the best recall@1 and mAP in \%. ft 90\& denotes the use of our dataset filtering method to refine the finetune dataset CUHK-PEDES, ultimately retaining 90\% of its data for training, while top50 is using a data filtering approach with a threshold set to top50 for selecting the pre-training dataset MALS.}
% \label{tab:method}
% \vspace{-3.1em}
% \end{table}
\subsection{Ablation Study and Further Discussion}
%In Table ~\ref{tab:pretrain}, we respectively compare the performance of our dataset filtering method applied to the CUHK-PEDES, RSTPReid and ICFG-PEDES datasets, in terms of pre-training Recall@1 and mAP metrics, alongside the final performance metrics of models that include finetuning on CUHK-PEDES, RSTPReid and ICFG-PEDES. 
\begin{table}[t]
\vspace{-.05in}
\fontsize{8}{8}\selectfont
    \setlength{\tabcolsep}{7pt}
    \renewcommand{\arraystretch}{1.0}
\scalebox{1.0}{\small
\begin{tabular}{c|cccc}
\toprule
Method   & top50 & ft 90\% & R@1    & mAP    \\ \midrule
Baseline* &       &        & 75.42 & 66.61 \\
Baseline* & \Checkmark    &        & 75.83 & 66.70 \\
Baseline* & \Checkmark     & \Checkmark      & 75.67 & 66.27 \\
\midrule
LoRA     & \Checkmark     &        & 74.40 & 64.95 \\
LoRA     & \Checkmark     & \Checkmark      & 74.29 (-1.38) & 65.59 (-0.68) \\
DoRA & \Checkmark     &                 & 75.49 & 66.92  \\ 
DoRA     & \Checkmark     & \Checkmark      & 75.73 (+0.06) & 66.75 (+0.48) \\ \midrule
WoRA (Ours)     & \Checkmark &  & 75.67 & 67.09  \\
WoRA (Ours)     & \Checkmark     & \Checkmark      & 76.38 (+0.71) & 67.22 (+0.95) \\ 
\bottomrule
\end{tabular}
}
\caption{Comparison of our WoRA with baseline, LoRA and DoRA in the different situations. Finally, the experiment shows that our methods achieve the best recall@1 and mAP in \%. ft 90\% denotes the utilization of our dataset filtering method to refine the finetune dataset CUHK-PEDES, ultimately retaining 90\% of its data for training, while top50 is using a data filtering approach with a threshold set to top50 for selecting the pre-training dataset MALS.}
\label{tab:method}
\vspace{-.35in}
\end{table}
% \vspace{-.25in}
\noindent\textbf{The impact of data filtering}. Here, ``baseline'' refers to the APTM method as implemented in our experimental setup, and "top50" denotes our data filtering approach with a threshold set to top50 for selecting the pre-training dataset MALS. From Table ~\ref{tab:pretrain}, it is evident that training the pre-trained models with our filtered dataset results in improvements of 6.26\% in Recall@1, demonstrating the efficacy of our dataset filtering for pre-training. Similarly, we find that filtering on the downstream dataset also facilitates the learning, since the filtering algorithm generally removes the noise. In Table ~\ref{tab:method}, the notation ``ft90\%'' signifies the leverage of our dataset filtering method to refine the finetune dataset CUHK-PEDES, ultimately retaining 90\% of its data for training. The model trained after applying ``ft90\%'' exhibits a 0.25\% increase in Recall@1, substantiating the effectiveness of dataset filtering on the comprehensive model.

\begin{figure}[t]
 \vspace{-.05in}
  \centering
    % \fbox{\rule{0pt}{2.5in} \rule{0.9\linewidth}{0pt}}
  \includegraphics[width=\linewidth]{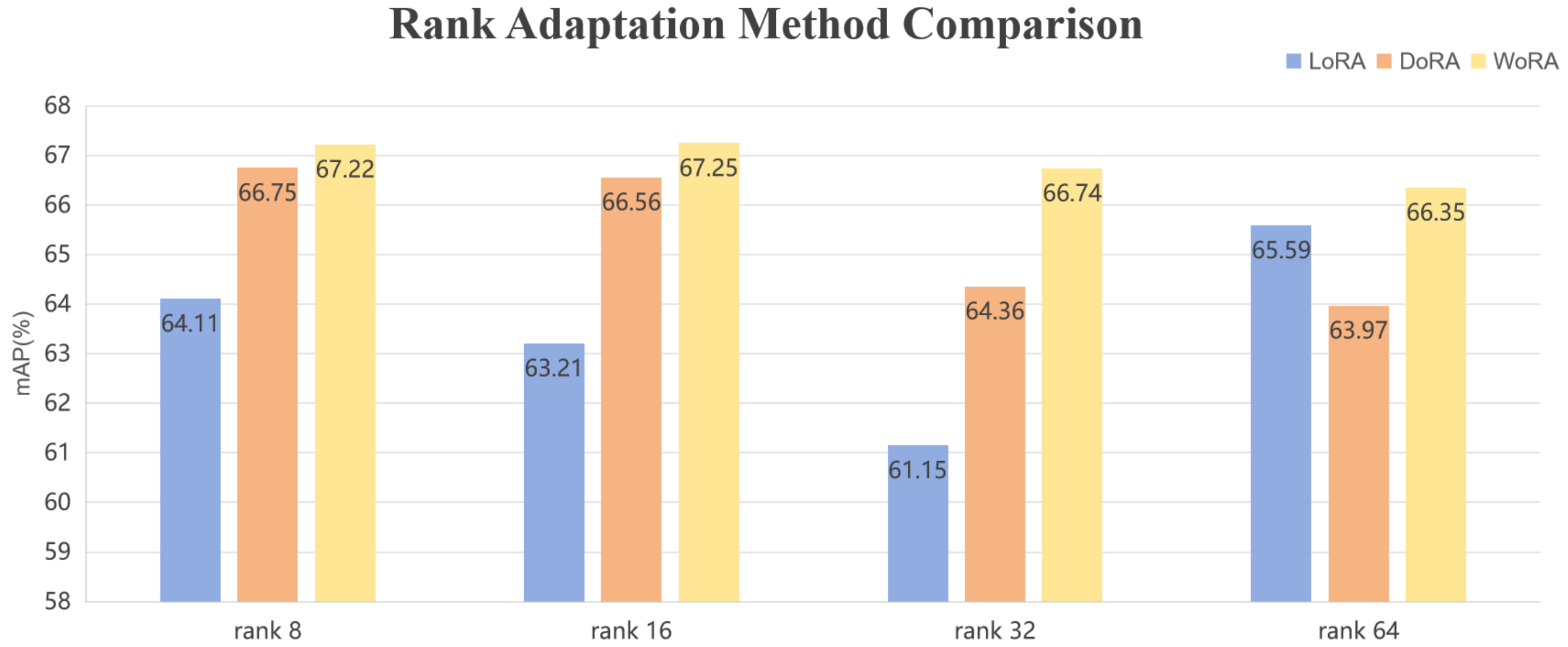}\vspace{-.15in}
  \caption{Compare the performance of LoRA, DoRA, and WoRA in terms of different rank. Our WoRA consistently surpasses both LoRA and DoRA on various rank settings. }
  \label{fig:8} 
    \vspace{-.21in}
\end{figure}

Moreover, we assess the impact of Weighted Low-Rank Adaptation (WoRA) on the complete model. Initially, comparing the impacts of using LoRA \cite{hu2021lora} and DoRA \cite{liu2024dora} models of Low-Rank Adaptation on our fully trained model revealed improvements in model speed but not in performance. Subsequently, applying WoRA to models trained on the top50 pre-training selection and the ft90\% finetuned dataset, we achieve a Recall@1 of 76.38\%, surpassing the baseline by 0.96\%, and an mAP of 67.22\%, exceeding the baseline by 0.61\%, thereby establishing state-of-the-art (SOTA) mAP performance. The computation time for the complete model is reduced from 23h to 18h, marking a 19.82\% acceleration. In Figure ~\ref{rank}, we compare the effects of different rank values on the performance of WoRA. The experiment shows that the model has the best performance on Recall@1,5,10 when rank is equal to 8. We can observe that model performance may not be as sensitive in different ranks. Performance can be improved basically with the WoRA model. The results we present in this paper are obtained using rank=8, which has achieved the best performance on Recall@1,5,10. As shown in Table ~\ref{tab:method}, we could observe two points (1) WoRA is better than both LoRA and DoRA, whether the performance of Recall@1 or mAP. %under the condition of using our top50 filtering strategy. 
(2) Furthermore, if we both leverage the top50 and ft90\% filtering strategy, WoRA is also the best performing, which surpasses the baseline by 0.71\% Recall@1 and 0.95\% mAP under the same conditions.
Meanwhile, in order to compare the effectiveness of WoRA. We control all the learning rates unchanged and compare the performance of LoRA, DoRA, and WoRA (ous) in Figure ~\ref{fig:8}. we use the same ranks for LoRA, DoRA, and WoRA for comparison. We can clearly see from the figure that under the condition of a fixed learning rate, the WoRA method proposed by us is consistently better than the LoRA and DoRA model mAP under the condition of the same rank. For example, when the same rank is 64, our method is +0.57\% mAP higher than LoRA. +2.38\% mAP higher than DoRA. Therefore, it can be clearly seen that our WoRA method is superior to DoRA and LoRA under the condition of a fixed learning rate.

\noindent\textbf{The impact of WoRA}. We further investigate the impact of Low-Rank Adaptation on pre-training, where ``WoRA'' in Table ~\ref{tab:pretrain} and  ~\ref{tab:method} signifies our Weighted Low-Rank Adaptation approach. Implementing WoRA in the pre-trained models led to a +6.72\% boost in Recall@1 and a +6.71\% enhancement in mAP, concurrently reducing the time by 22.22\%, thereby validating the efficiency of our dataset filtering and WoRA in model pre-training.

\section{Conclusion}
In this work, we introduce a new Filtering-WoRA paradigm, which contains a filtering algorithm to identify this crucial data subset and WoRA layers (Weighted Low-Rank Adaptation) for light fine-tuning. Filtering strategy for image-text pairs within language-based person search datasets, designed to isolate a core set from large-scale, noise-containing datasets of generated image-text pairs. WoRA (Weighted Low-Rank Adaptation) learning strategy to efficiently update the portion of model parameters. Extensive experiments indicate that our approach is 19.82\% faster than existing language-based person search methods while maintaining comparable accuracy with state-of-the-art (SOTA) language-based person search models. On three public benchmarks, CUHK-PEDES, RSTPReeid, and ICFG-PEDES, our method achieves competitive recall rates and mean Average Precision (mAP). 

\section*{Acknowledgements}
This work is supported by the University of Macau Start-up Research Grant SRG2024-00002-FST and Multi-Year Research Grant MYRG-GRG2024-00077-FST-UMDF, and National Social Science Foundation Major Project in Art (No.2024ZDE054).

%We hope our contributions can advance the community's efforts in unified text-based person search.

% In this work, we introduce a new Filtering-WoRA paradigm, which contains a filtering algorithm to identify this crucial data subset and WoRA layers (Weighted Low-Rank Adaptation) for lite fine-tuning. Filtering strategy for image-text pairs within language-based person search datasets, designed to isolate a core set from large-scale, noise-containing datasets of generated image-text pairs. WoRA (Weighted Low-Rank Adaptation) learning strategy to efficiently updating portion of model parameters. Extensive experiments indicate that our approach is 19.82\% faster than existing language-based person search methods, while maintaining comparable accuracy with state-of-the-art (SOTA) language-based person search models. On three public benchmarks, CUHK-PEDES, RSTPReeid and ICFG-PEDES, our method achieves competitive recall rates and mean Average Precision (mAP). We hope our contributions can advance the community's efforts in unified text-based person search.

\bibliographystyle{ACM-Reference-Format}
\bibliography{sample-base}

% \appendix









\end{document}